\documentclass[5p,times,authoryear]{elsarticle}

\usepackage{epsfig}
\usepackage{graphicx}

\usepackage{amssymb}
\usepackage{amsmath}

\setcitestyle{square,numbers}
\usepackage{setspace} 

\usepackage{xcolor}
\usepackage{amsthm}
\usepackage{mathrsfs} 
\usepackage{tabularx}
\usepackage{float}
\usepackage{times}
\usepackage{adjustbox}

\def\V{\mathbb{V}}

\def\G{\mathcal{G}}
\def\S{\mathbb{S}}
\def\RR{\mathscr{R}}

\newcommand{\Sk}{\hat{\mathbb{S}}^{(k)}}
\newcommand{\Vk}{\hat{\mathbb{V}}^{(k)}}

\begin{document}

\begin{frontmatter}

\title{A sliced-Wasserstein distance-based approach for out-of-class-distribution detection}

\author[affil_lab,affil_bme]{Mohammad Shifat-E-Rabbi\corref{cor1}} % \fnref{label2}
\ead{mr2kz@virginia.edu}
%\ead[url]{www.cea-ifac.es}

\author[affil_lab,affil_ece]{Abu Hasnat Mohammad Rubaiyat}
\ead{ar3fx@virginia.edu}

\author[affil_lab,affil_ece]{Yan Zhuang}
\ead{yz8bk@virginia.edu}
%\ead[url]{www2.cea-ifac.es}

\author[affil_lab,affil_bme,affil_ece]{Gustavo K. Rohde}
\ead{gustavo@virginia.edu}

%\fntext[label2]{Nota al pie para el autor 1}
\cortext[cor1]{Corresponding author.}

\address[affil_lab]{Imaging and Data Science Laboratory, University of Virginia, Charlottesville, VA 22908, USA}
\address[affil_bme]{Department of Biomedical Engineering, University of Virginia, Charlottesville, VA 22908, USA}
%\address[Second]{Departamento de Automatica, Ingenieraa Electranica e Informatica, Universidad Politacnica de Madrid,  C/ Josa Gutiarrez Abascal, na2, 28006, Madrid,  Espaaa.}
\address[affil_ece]{Department of Electrical and Computer Engineering, University of Virginia, Charlottesville, VA 22908, USA}

\begin{abstract}
There exist growing interests in intelligent systems for numerous medical imaging, image processing, and computer vision applications, such as face recognition, medical diagnosis, character recognition, and self-driving cars, among others. These applications usually require solving complex classification problems involving complex images with unknown data generative processes. In addition to recent successes of the current classification approaches relying on feature engineering and deep learning, several shortcomings of them, such as the lack of robustness, generalizability, and interpretability, have also been observed. These methods often require extensive training data, are computationally expensive, and are vulnerable to out-of-distribution samples, e.g., adversarial attacks. Recently, an accurate, data-efficient, computationally efficient, and robust transport-based classification approach has been proposed, which describes a generative model-based problem formulation and closed-form solution for a specific category of classification problems. However, all these approaches lack mechanisms to detect test samples outside the class distributions used during training. In real-world settings, where the collected training samples are unable to exhaust or cover all classes, the traditional classification schemes are unable to handle the unseen classes effectively, which is especially an important issue for safety-critical systems, such as self-driving and medical imaging diagnosis. In this work, we propose a method for detecting out-of-class distributions based on the distribution of sliced-Wasserstein distance from the Radon Cumulative Distribution Transform (R-CDT) subspace. We tested our method on the MNIST and two medical image datasets and reported better accuracy than the state-of-the-art methods without an out-of-class distribution detection procedure.
\end{abstract}


\end{frontmatter}

% \linenumbers
\onecolumn

\section*{Supplemental information}

\subsection*{Description of purpose}
There exists a growing need for intelligent systems for numerous medical imaging, image processing, and computer vision applications, such as face recognition, medical diagnosis (pathology, radiology), optical character recognition, and self-driving cars, among others \cite{lecun2015deep,boland2001neural}. In most cases, these applications require solving complex classification problems involving complex object images whose underlying data generative processes are usually unknown. Current approaches to solving these classification problems rely on generic feature approximation and regression approaches, such as feature engineering and deep learning \cite{lecun2015deep,he2016deep}. Recently, a new transport-based modeling approach has been proposed, which describes a generative model-based problem formulation and solution to a specific category of classification problems in closed-form \cite{shifat2021radon,kolouri2015radon}. However, these approaches lack mechanisms to detect test samples outside the class distributions used during the training process. In real-world settings, where the collected training samples are unable to exhaust or cover all classes, the traditional classification schemes are unable to handle the unseen classes effectively \cite{geng2020recent}. This is especially an important issue for safety-critical systems, such as self-driving \cite{shaheen2022continual} and medical imaging diagnosis \cite{guan2021domain}. In this work, we propose a method for detecting out-of-class distributions based on the distribution of sliced-Wasserstein distance from the Radon Cumulative Distribution Transform (R-CDT) subspaces.

\subsection*{Method}
We begin with a generative model-based problem statement for the type of classification problems we consider here. As mentioned in \cite{shifat2021radon}, in many image classification problems, image patterns can be thought of being instances of a certain prototype (or template) observed under some unknown spatial deformation patterns. The generative model stated below formalizes this statement.\\\\
{\bf{Generative model:}}
{\em Let $\G$ be the set of spatial deformations. The mass (image intensity) preserving generative model for the $k^{\mbox{th}}$ image class is defined to be the set}
\begin{equation}
\mathbb{S}^{(k)}=\left\{s_j^{(k)}|s_j^{(k)}=\RR^{-1}\left(\left({g_j^\theta}\right)^\prime\widetilde{\varphi}^{(k)}\circ g^\theta_j\right), \forall g^\theta_j\in\G \right\}.
\label{eq:2dgenerative_model}
\end{equation}
The notation $s_j^{(k)}$ here is meant to denote the $j^{\text{th}}$ image from the $k^{\text{th}}$ class, $g^\theta_j$ denotes the deformation, $\varphi^{(k)}$ denotes the template image pattern, $\widetilde{\varphi}^{(k)}$ denotes the template in the Radon transform space, $\RR^{-1}(\cdot)$ denotes the inverse Radon transform operator, and `$\circ$' denotes the functional composition operator. We are now ready to define a mathematical description for a generative model-based problem statement using the definition above.\\\\
{\bf{Classification problem:}}
{\em Let $\G$ be the set of deformations, and let $\mathbb{S}^{(k)}$ be defined as in equation~\eqref{eq:2dgenerative_model}. Given training samples $\{s^{(1)}_1, s^{(1)}_2, \cdots\}$ (class 1), $\{s^{(2)}_1, s^{(2)}_2,$ $\cdots\}$ (class 2), $\cdots$ as training data, determine the class $(k)$ of an unknown image $s$.}
\begin{figure}[h]
	\centering
	\includegraphics[width=0.9\textwidth]{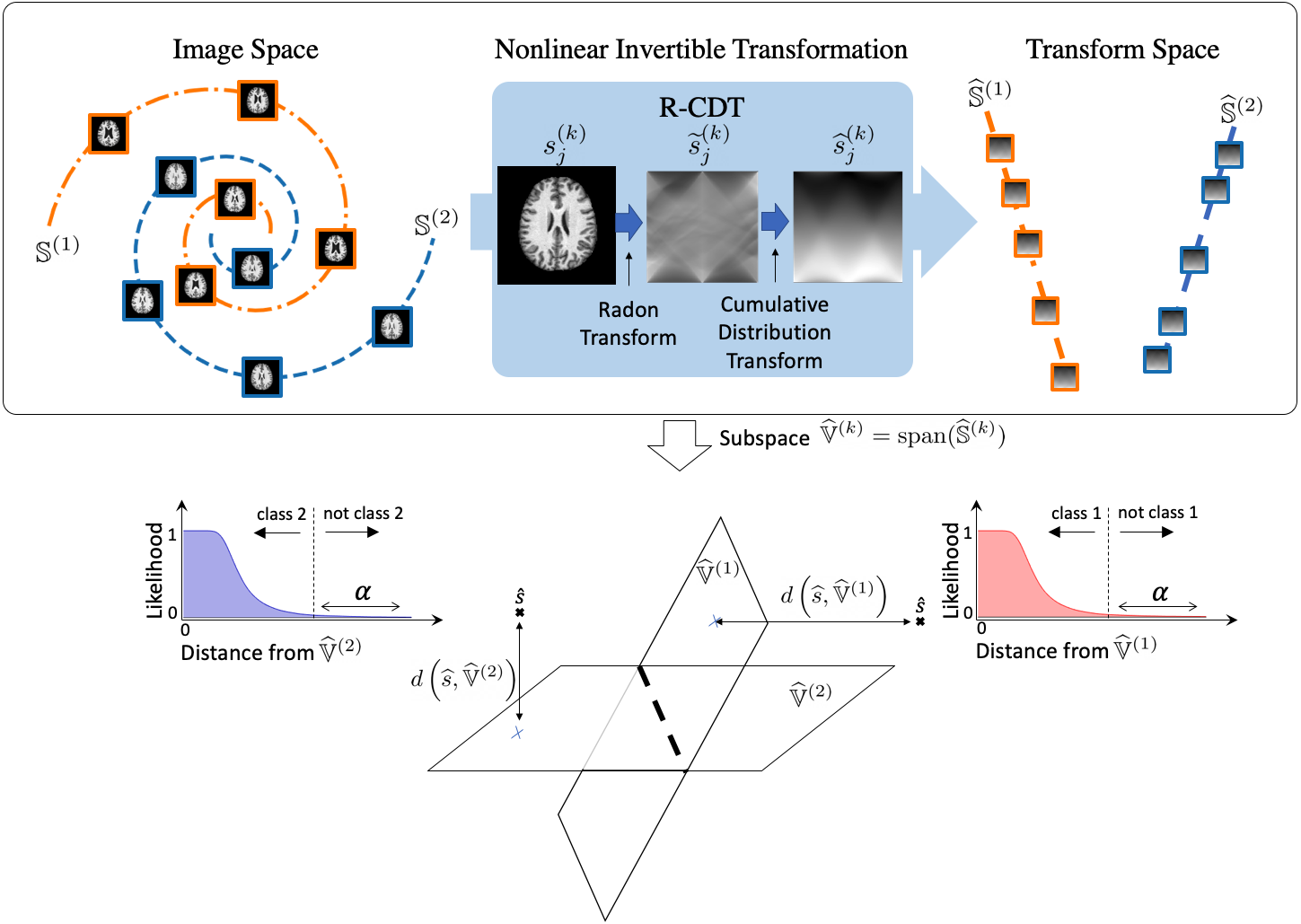}
	\caption{Conceptual system diagram outlining the proposed out-of-class distribution detection method.}
	\label{fig1}
\end{figure}

It is important to note that the generative model above yields nonconvex (and hence nonlinear) signal classes, and thereby complicates the classification problem. Next, we discuss the solution to the classification problem.\\\\
{\bf{Solution and new contribution:}}
As mentioned in \cite{shifat2021radon}, the R-CDT transform \cite{kolouri2015radon} can be used to drastically simplify the solution to the classification problem above. The R-CDT space generative model can be expressed as
\begin{eqnarray}
\widehat{\mathbb{S}}^{(k)}&=&\left\{\widehat{s}_j^{(k)}|\widehat{s}_j^{(k)}={\left(g_j^\theta\right)}^{-1}\circ \widehat{\varphi}^{(k)}, \forall g_j^\theta\in\G \right\}. 
\end{eqnarray}
Under some reasonable assumptions (see \cite{shifat2021radon}), the R-CDT transform simplifies the data geometry in a way that image classes become convex, and thereby simplifies the classification problem. In fact, according to \cite{ shifat2021radon}, the convex set $\Sk$ can be approximated as a subspace $\Vk$ as $\widehat{\V}^{(k)}=\mbox{span}\left(\Sk\right)$. 

The training algorithm consists of estimating the subspace $\Vk$ corresponding to the transform space $\Sk$ given sample data $\{s_1^{(k)},$ $s_2^{(k)},$ $\cdots\}$. Naturally, the first step is to transform the training data to obtain $\{\widehat{s}_1^{(k)}, \widehat{s}_2^{(k)}, \cdots \}$. Then the subspace $\Vk$ is approximated as $\Vk=\text{span}\left\{ \widehat{s}_1^{(k)}, \widehat{s}_2^{(k)}, \cdots  \right \}$ \cite{shifat2021radon}. As far as a test procedure for determining the class of some unknown signal or image $s$, it suffices to measure the distance between $\widehat{s}$ and the nearest point in each subspace $\Vk$ corresponding to the generative model $\widehat{\S}^{(k)}$. In other words, the class of the unknown sample can be decoded by solving $\arg \min_k d^2(\widehat{s},\Vk)$ \cite{shifat2021radon}.

However, the method in \cite{shifat2021radon} does not contain a procedure to detect samples outside the classes used in the training process. %In this work, we extend the method in \cite{shifat2021radon} and incorporate an out-of-class-detection procedure in it. In that goal, we obtained an empirical likelihood function as a function of distance from the subspaces. 
In this paper, we incorporate an out-of-class-distribution detection procedure by empirically estimating a likelihood function based on the distances from the subspace corresponding to a class.
%First, we obtained the univariate probability density functions (PDF) $p_{k}(x)$ over $x$ using the empirical distance measurements of the validation set from the trained subspace of each class.
First, we obtain a probability density function of distances for each class based on the distance measurements of a validation set from the corresponding subspace. Let the probability density function is denoted as $p_{k}(x)$, where $x$ denotes the distance of an image sample $s$ from the $k$-th subspace $\Vk$ in the R-CDT space, i.e., $d\left(\widehat{s},\Vk\right)$.
%Here, $x$ denotes the distance of an image sample from the $k$-th subspace $\Vk$ in the R-CDT space. 
A kernel density estimation technique was employed to obtain the PDFs from the distance measurements. The empirical likelihood for the k-th class is then computed as
\begin{align}
    \mathcal{L}^{(k)}(x)=1-\int_{-\infty}^{x}p_k(u)du
\end{align}
Once we obtain the likelihood functions, a particular confidence interval is calculated, based on which we can accept or reject a test sample to belong to a particular class. Fig.~\ref{fig1} shows an outline of the proposed method.

\subsection*{Results}
We compared the proposed method with conventional classification methods with respect to classification accuracy. In this respect, we have identified four state-of-the-art methods: MNISTnet \cite{paszke2019pytorch} (a shallow CNN model based on PyTorch's official example), the standard VGG11 model \cite{simonyan2014very}, the standard Resnet18 model \cite{he2016deep}, the standard k-nearest neighbors (kNN) classifier model \cite{kramer2013dimensionality}, and the R-CDT NS method \cite{shifat2021radon}. All methods saw the same set of training and test images.
\begin{figure}[!hbt]
	\centering
	\includegraphics[width=1.0\textwidth]{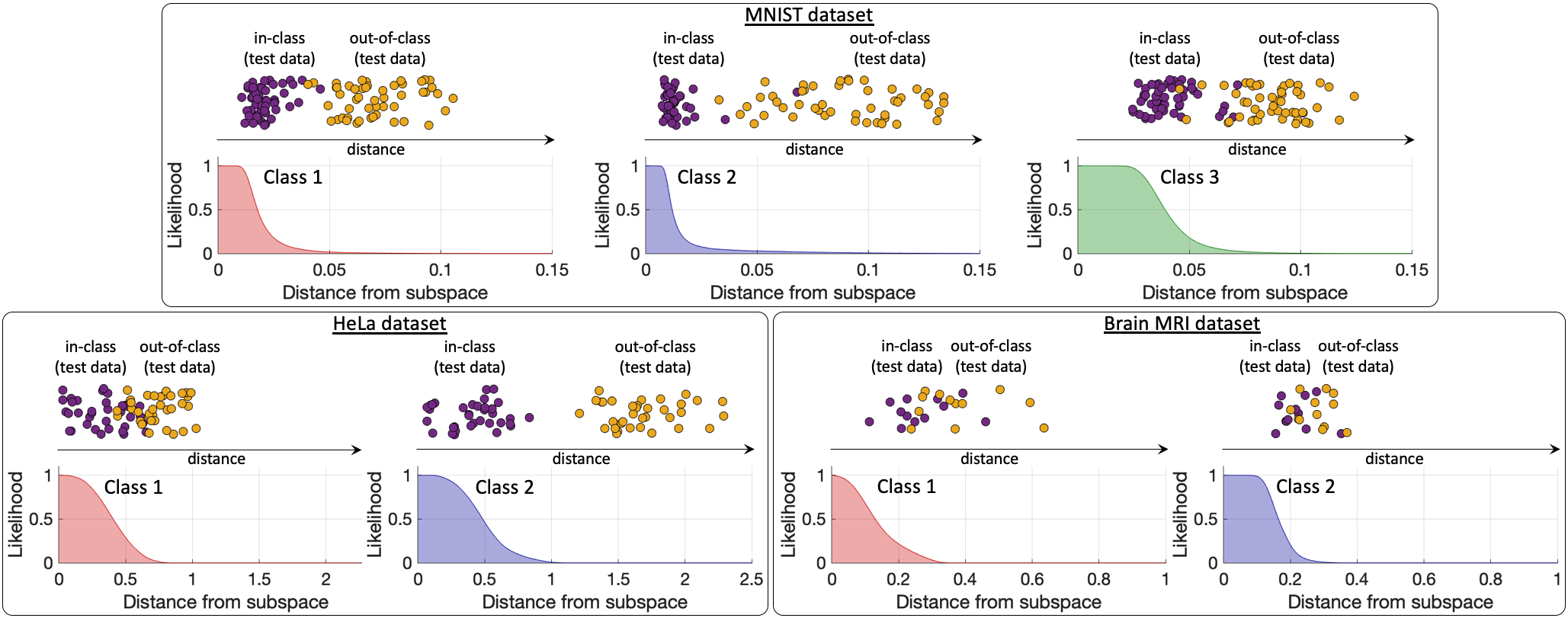}
	\caption{Estimated likelihood functions and the scatter plots of sampled in-class and out-of-class test data.}
	\label{fig2}
\end{figure}

Fig.~\ref{fig2} shows the estimated likelihood functions for different classes of a data and the scatter plots of a few sampled in-class and out-of-class test data. We can see clear separations between the in-class and out-of-class test data along the distance axis. The comparative test accuracy of different methods on three datasets are shown in Table~\ref{tab1}. The proposed method was implemented with different confidence intervals $\alpha$ (see Table~\ref{tab1} and Fig.~\ref{fig1}). A collection of classes in the datasets were randomly chosen as the in-class data and one class was randomly chosen as the out-of-class data. The proposed method outperforms the other methods, which are designed without the out-of-class detection procedure.

\begin{table}[!hbt]
\caption{Test accuracy (\%) of the methods.}
\begin{adjustbox}{width=\columnwidth,center}
\label{tab1}
\begin{tabular}{lccccc|ccc}
\hline
      & \multicolumn{5}{c|}{Methods without out-of-class detection procedure} & \multicolumn{3}{c}{Proposed method}                                                                                                                                                                                    \\ \cline{2-9} 
      & MNISTnet \cite{paszke2019pytorch}      & VGG11 \cite{simonyan2014very}      & Resnet18 \cite{he2016deep}      & k-NN \cite{kramer2013dimensionality}     & R-CDT NS \cite{shifat2021radon}     & \begin{tabular}[c]{@{}c@{}}Proposed\\ ($\alpha=1\%$)\end{tabular} & \multicolumn{1}{c}{\begin{tabular}[c]{@{}c@{}}Proposed\\ ($\alpha=5\%$)\end{tabular}} & \multicolumn{1}{c}{\begin{tabular}[c]{@{}c@{}}Proposed\\ ($\alpha=10\%$)\end{tabular}} \\ \hline
MNIST \cite{lecun1998gradient}     & $74.64$ & $-$ & $74.80$ & $73.59$ & $74.71$ & $86.74$ & $\bf{95.44}$ & $92.69$      \\
HeLa cell \cite{boland2001neural} & $45.78$ & $41.56$ & $40.94$ & $41.64$ & $63.05$ & $71.41$ & $76.48$      & $\bf{79.45}$ \\
Brain MRI \cite{fu2021automated} & $16.99$ & $18.71$ & $16.97$ & $14.43$ & $11.94$ & $54.49$ & $62.77$      & $\bf{66.77}$
\\ \hline
\end{tabular}
\end{adjustbox}
\end{table}

\subsection*{Conclusions}
This work proposes an enhanced end-to-end classification system with a mechanism to detect samples outside the class distributions used in training. The proposed method is pertinent to a specific category of image classification problems where image classes can be thought of being an instance of a template observed under a set of spatial deformations. The method is mathematically coherent, understandable, non-iterative, requires no hyper-parameters to tune, and is simple enough to be implemented without GPU support. We obtain superior performance by expanding upon the recently published R-CDT NS classification method \cite{shifat2021radon}, which can also be interpreted as the `nearest' sliced-Wasserstein distance method. The proposed mathematical solution attains high classification accuracy (compared with state-of-the-art end-to-end systems without out-of-class detection procedure) on the MNIST and two medical image datasets.

\subsection*{Is the work published elsewhere?}
The method we propose here is an extension of the method proposed in \cite{shifat2021radon}. The empirical likelihood estimation based classification method as shown in Fig.~\ref{fig1} and the new results in Fig.~\ref{fig2} and Table~\ref{tab1} have not been published anywhere.

\section*{Acknowledgments}
This work was supported in part by NIH grant GM130825, and NSF grant 1759802.

{\small
\bibliographystyle{elsarticle-num}
\bibliography{egbib}
}

\end{document}